\documentclass[runningheads]{llncs}

\usepackage{graphicx}
\usepackage{amsmath,amssymb}
\usepackage{cite}
\usepackage{xcolor}
\usepackage{bbding}
\usepackage{multirow}
\usepackage{subcaption}
\usepackage{enumitem}

\newcommand{\blue}[1]{\textcolor{blue}{#1}}
\newcommand{\red}[1]{\textcolor{red}{#1}}

\newcommand\Tstrut{\rule{0pt}{2.5ex}}
\newcommand\Bstrut{\rule[-1ex]{0pt}{0pt}}

\def\HS{\hspace{\fontdimen2\font}}
\def\HSE{\HS\HS\HS\HS\HS\HS\HS\HS}

\begin{document}

\title{Fast, Accurate, and Lightweight Super-Resolution\\with Cascading Residual Network}

\titlerunning{Fast, Accurate, and Lightweight Super-Resolution with CARN}
\authorrunning{Namhyuk Ahn, Byungkon Kang, and Kyung-Ah Sohn}

\author{Namhyuk Ahn\inst{}\orcidID{0000-0003-1990-9516} \and
        Byungkon Kang\inst{}\orcidID{0000-0001-8541-2861} \and
        Kyung-Ah Sohn\inst{}\orcidID{0000-0001-8941-1188}}
\institute{Department of Computer Engineering,\\ Ajou University\\
\email{\{aa0dfg,byungkon,kasohn\}@ajou.ac.kr}}

\maketitle

\begin{abstract}
In recent years, deep learning methods have been successfully applied to single-image super-resolution tasks. Despite their great performances, deep learning methods cannot be easily applied to real-world applications due to the requirement of heavy computation. In this paper, we address this issue by proposing an accurate and lightweight deep network for image super-resolution. In detail, we design an architecture that implements a \textit{cascading mechanism} upon a residual network. We also present variant models of the proposed cascading residual network to further improve efficiency. Our extensive experiments show that even with much fewer parameters and operations, our models achieve performance comparable to that of state-of-the-art methods.
\end{abstract}

\section{Introduction}
\label{sec:intro}
Super-resolution (SR) is a computer vision task that reconstructs a high-resolution (HR) image from a low-resolution (LR) image. Specifically, we are concerned with single image super-resolution (SISR), which performs SR using a single LR image. SISR is generally difficult to achieve due to the fact that computing the HR image from an LR image is a many-to-one mapping. Despite such difficulty, SISR is a very active area because it can offer the promise of overcoming resolution limitations, and could be used in a variety of applications such as video streaming or surveillance system.

\begin{figure}[t]
\centering
\includegraphics[width=\textwidth]{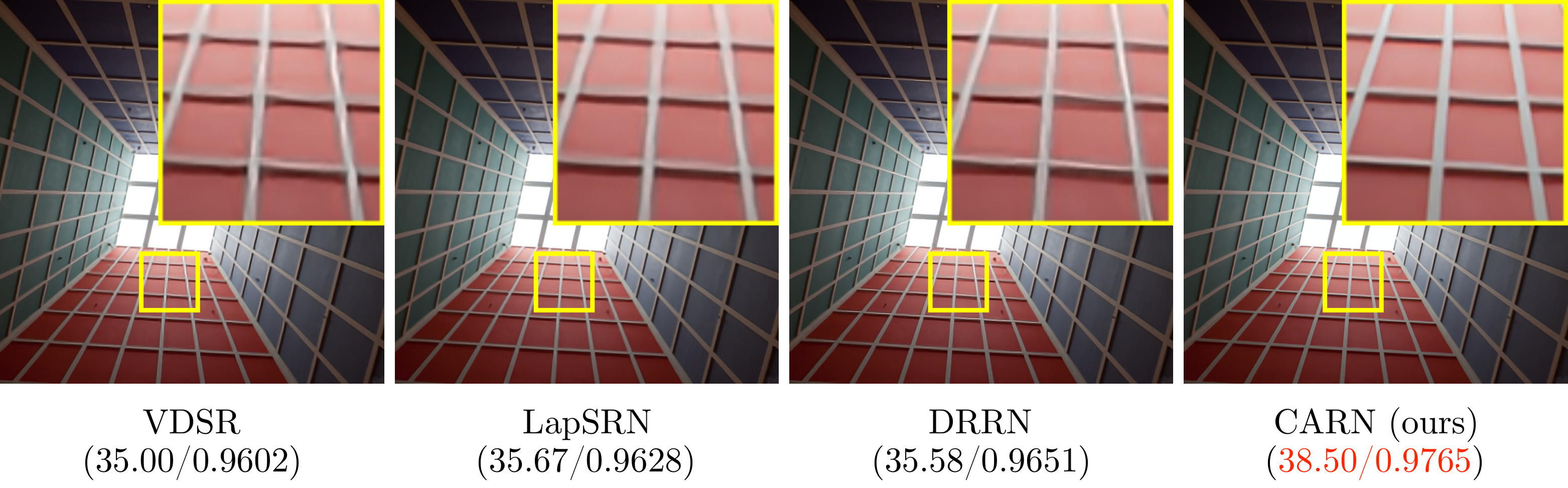}
\caption{Super-resolution result of our methods compared with existing methods.}
\label{fig:intro}
\end{figure}

Recently, convolutional neural network-based(CNN-based) methods have provided outstanding performance in SISR tasks\cite{srcnn2014,vdsr2016,lapsrn2017}. From the SRCNN\cite{srcnn2014} that has three convolutional layers to MDSR\cite{mdsr2017} that has more than 160 layers, the depth of the network and the overall performance have dramatically grown over time. However, even though deep learning methods increase the quality of the SR images, they are not suitable for real-world scenarios. From this point of view, it is important to design lightweight deep learning models that are practical for real-world applications. One way to build a lean model is reducing the number of parameters. There are many ways to achieve this\cite{han2015deep,squeezenet}, but the most simple and effective approach is to use a \textit{recursive network}. For example, DRCN\cite{drcn2016} uses a recursive network to reduce redundant parameters, and DRRN\cite{drnn2017} improves DRCN by adding a residual architecture to it. These models decrease the number of model parameters effectively when compared to the standard CNN and show good performance. However, there are two downsides to these models: 1) They first upsample the input image before feeding it to the CNN model, and 2) they increase the depth or the width of the network to compensate for the loss due to using a recursive network. These points enable the model to maintain the details of the image when reconstructed, but at the expense of the increased number of operations and inference time.

Most of the works that aim to build a lean model focused primarily on reducing the number of parameters. However, as mentioned above, the number of operations is also an important factor to consider in real-world scenarios. Consider a situation where an SR system operates on a mobile device. Then, the execution speed of the system is also of crucial importance from a user-experience perspective. Especially the battery capacity, which is heavily dependent on the amount of computation performed, becomes a major problem. In this respect, reducing the number of operations in the deep learning architectures is a challenging and necessary step that has largely been ignored until now. Another scenario relates to applying SR methods to video streaming services. The demand for streaming media has skyrocketed and hence requires large storage to store massive multimedia data. It is therefore imperative to compress data using lossy compression techniques before storing. Then, an SR technique can be applied to restore the data to the original resolution. However, because latency is the most critical factor in streaming services, the decompression process (i.e., super-resolution) has to be performed in real-time. To do so, it is essential to make the SR methods lightweight in terms of the number of operations.

To handle these requirements and improve the recent models, we propose a Cascading residual network (CARN) and its variant CARN-Mobile (CARN-M). We first build our CARN model to increase the performance and extend it to CARN-M to optimize it for speed and the number of operations. Following the FSRCNN\cite{fsrcnn2016}, CARN family take the LR images and compute the HR  counterparts as the output of the network. The middle parts of our models are designed based on the ResNet\cite{resnet}. The ResNet architecture has been widely used in deep learning-based SR methods\cite{drnn2017,mdsr2017} because of the ease of training and superior performance. In addition to the ResNet architecture, CARN uses a \textit{cascading mechanism} at both the local and the global level to incorporate the features from multiple layers. This has the effect of reflecting various levels of input representations in order to receive more information. In addition to the CARN model, we also provide the CARN-M model that allows the designer to tune the trade-off between the performance and the \textit{heaviness} of the model. It does so by means of the efficient residual block (residual-E) and recursive network architecture, which we describe in more detail in Section~\ref{sec:method}.

In summary, our main contributions are as follows: \textbf{1)} We propose CARN, a neural network based on the cascading modules, which achieves high performance on SR task (Fig.~\ref{fig:intro}). Our cascading modules, effectively boost the performance via multi-level representation and multiple shortcut connections. \textbf{2)} We also propose CARN-M for efficient SR by combining the efficient residual block and the recursive network scheme. \textbf{3)} We show through extensive experiments, that our model uses only a modest number of operations and parameters to achieve competitive results. Our CARN-M, which is the more lightweight SR model, shows comparable results to others with much fewer operations.

\section{Related Work}
\label{sec:related_work}
Since the success of AlexNet~\cite{alexnet} in image recognition task~\cite{imagenet2009}, many deep learning approaches have been applied to diverse computer vision tasks~\cite{ssd,fasterrcnn,deconvnet,color}. The SISR task is one such task, and we present an overview of the deep learning-based SISR in section~\ref{subsec:sisr}. Another area we deal with in this paper is model compression. Recent deep learning models focus on squeezing model parameters and operations for application in low-power computing devices, which has many practical benefits in real-world applications. We briefly review in section~\ref{subsec:efficient_nn}.

\subsection{Deep Learning Based Image Super-Resolution}
\label{subsec:sisr}
Recently, deep learning based models have shown dramatic improvements in the SISR task. Dong et al.~\cite{srcnn2014} first proposed a deep learning-based SR method, SRCNN, which outperformed traditional algorithms. However, SRCNN has a large number of operations compared to its depth, since network takes upsampled images as input. Taking a different approach from SRCNN, FSRCNN~\cite{fsrcnn2016} and ESPCN~\cite{espcn2016} upsample images at the end of the networks. By doing so, it leads to the reduction in the number of operations compared to the SRCNN. However, the overall performance could be degraded if there are not enough layers after the upsampling process. Moreover, they cannot manage multi-scale training, as the input image size differs for each upsampling scale.

Despite the fact that the power of deep learning comes from \textit{deep} layers, the aforementioned methods have settled for shallow layers because of the difficulty in training. To better harness the depth of deep learning models, Kim et al.~\cite{vdsr2016} proposed VDSR, which uses \textit{residual learning} to map the LR images $\textbf{x}$ to their residual images $\textbf{r}$. Then, VDSR produces the SR images $\textbf{y}$ by adding the residual back into the original, \textit{i.e.}, $\mathbf{y = x + r}$. On the other hand, LapSRN~\cite{lapsrn2017} uses a Laplacian pyramid architecture to increase the image size gradually. By doing so, LapSRN effectively performs SR on extremely low-resolution cases with a fewer number of operations compared to VDSR.

Another issue of deep learning-based SR is how to reduce the parameters and operation. For example, DRCN~\cite{drcn2016} uses a recursive network to reduce parameters by engaging in redundant usages of a small number of parameters. DRRN~\cite{drnn2017} improves  DRCN by combining the recursive and residual network schemes to achieve better performance with fewer parameters. However, DRCN and DRRN use very deep networks to compensate for the loss of performance and hence these require heavy computing resources. Hence, we aim to build a model that is lightweight in both size and computation. We will briefly discuss previous works that address such model efficiency issues in the following section.

\begin{figure}[t]
    \centering
    \includegraphics[width=\textwidth]{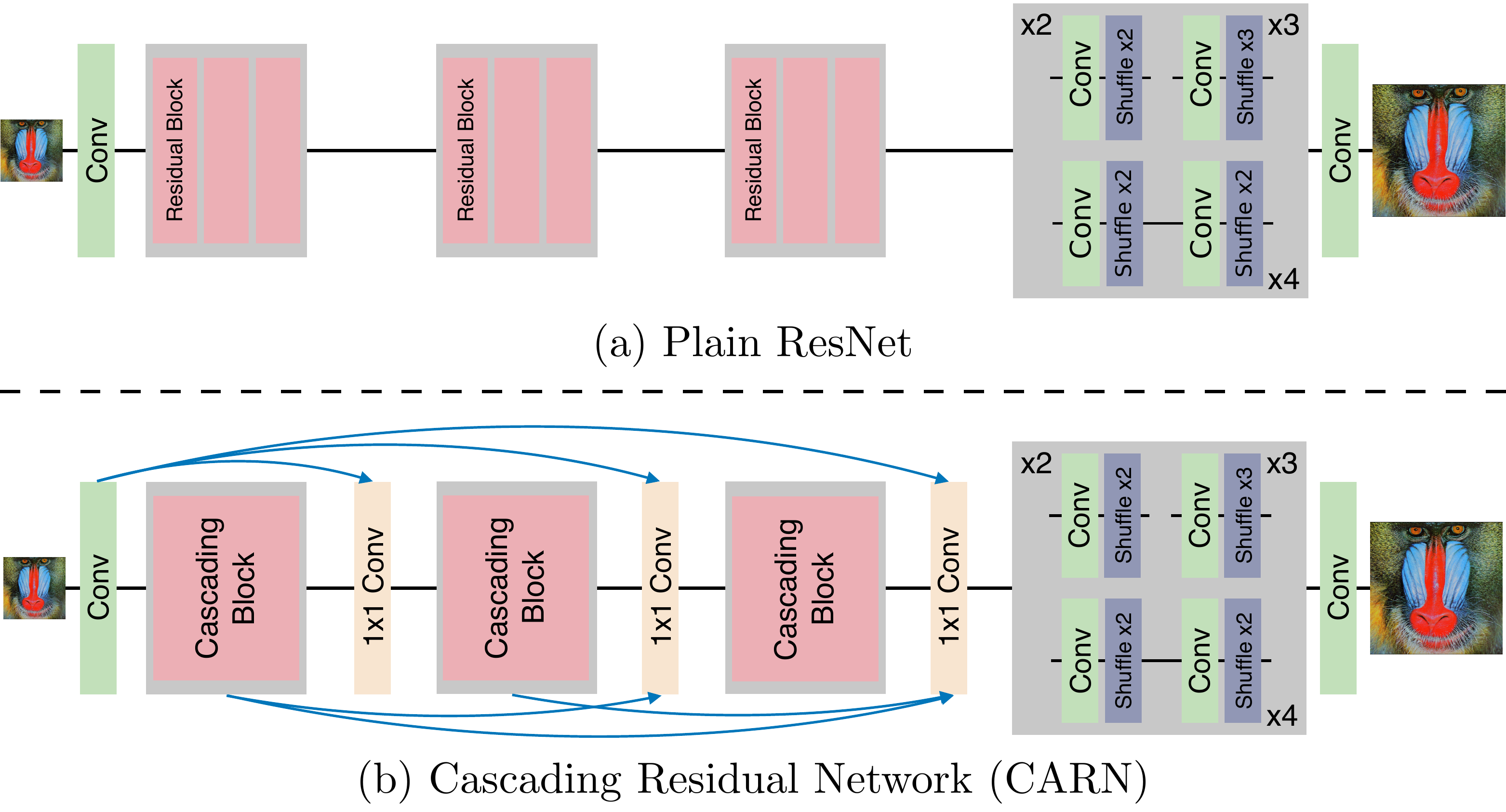}
    \caption{Network architectures of plain ResNet (\textbf{top}) and the proposed CARN (\textbf{bottom}). Both models are given an LR image and upsample to HR at the end of the network. In the CARN model, each residual block is changed to a cascading block. The blue arrows indicate the global cascading connection.}
    \label{fig:model}
\end{figure}

\subsection{Efficient Neural Network}
\label{subsec:efficient_nn}
There has been rising interest in building small and efficient neural networks~\cite{squeezenet,han2015deep,mobilenets}. These approaches can be categorized into two groups: \textbf{1)} Compressing pretrained networks, and \textbf{2)} designing small but efficient models. Han et al.~\cite{han2015deep} proposed deep compressing techniques, which consist of pruning, vector quantization, and Huffman coding to reduce the size of a pretrained network. In the latter category, SqueezeNet~\cite{squeezenet} builds an AlexNet-based architecture and achieves comparable performance level with 50$\times$ fewer parameters. MobileNet~\cite{mobilenets} builds an efficient network by applying depthwise separable convolution introduced in Sifre et al.~\cite{dwconv}. Because of this simplicity, we also apply this technique in the residual block with some modification to achieve a lean neural network.

\section{Proposed Method}
\label{sec:method}
As mentioned in Section~\ref{sec:intro}, we propose two main models: CARN and CARN-M. CARN is designed to be a high-performing SR model while suppressing the number of operations compared to the state-of-the-art methods. Based on CARN, we design CARN-M, which is a much more efficient SR model in terms of both parameters and operations.

\subsection{Cascading Residual Network}
\label{subsec:carn}
Our CARN model is based on ResNet~\cite{resnet2016}. The main difference between CARN and ResNet is the presence of local and global cascading modules. Fig.~\ref{fig:model} (b) graphically depicts how the global cascading occurs. The outputs of intermediary layers are \textit{cascaded} into the higher layers, and finally converge on a single 1x1 convolution layer. Note that the intermediary layers are implemented as cascading blocks, which host local cascading connections themselves. Such local cascading operations are shown in Figure~\ref{fig:model}(c) and (d). Local cascading is almost identical to a global one, except that the unit blocks are plain residual blocks.

\begin{figure}[tbp]
    \centering
    \includegraphics[width=\textwidth]{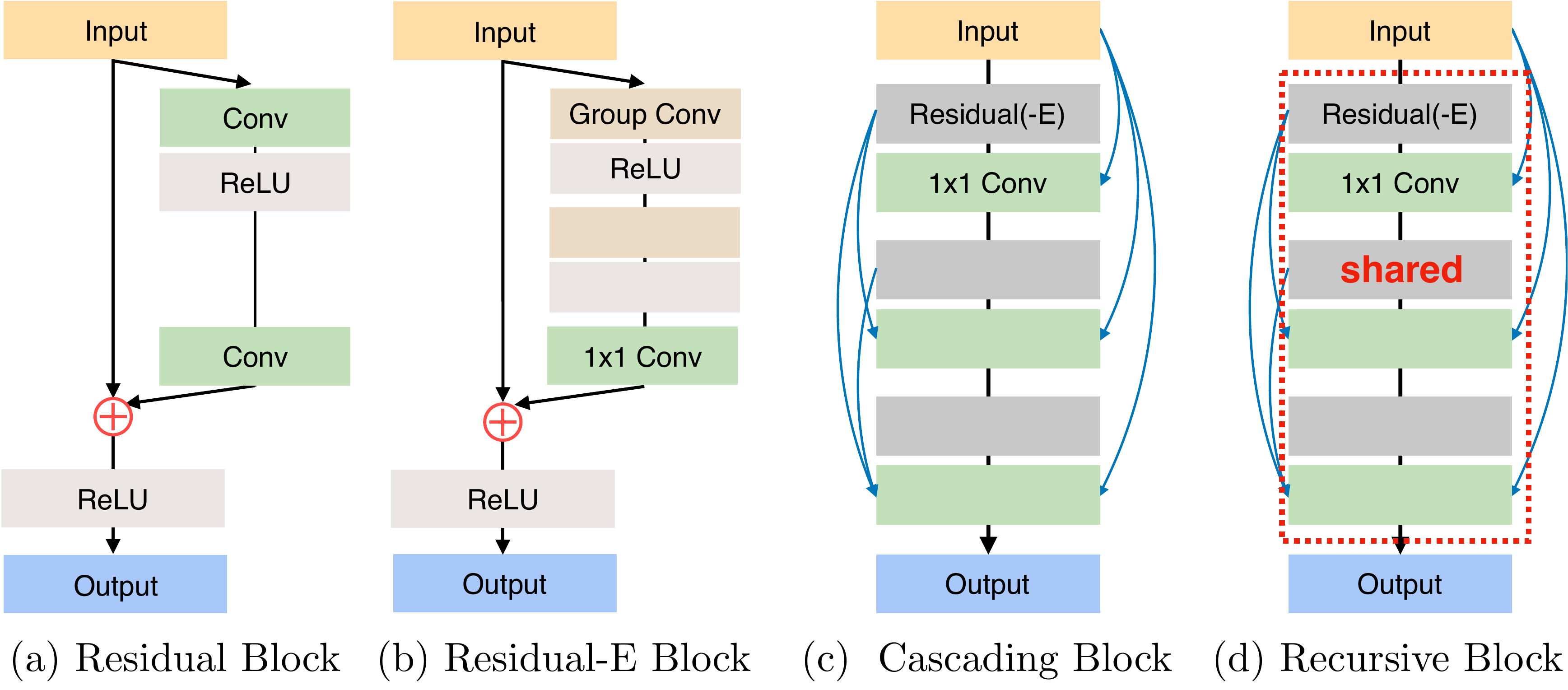}
    \caption{Simplified structures of (a) residual block (b) efficient residual block (residual-E), (c) cascading block and (d) recursive cascading block. The \red{$\oplus$} operations in (a) and (b) are element-wise addition for residual learning.}
    \label{fig:block}
\end{figure}

To express the implementation formally, let $f$ be a convolution function and $\tau$ be an activation function. Then, we can define the $i$-th residual block $R_i$, which has two convolutions followed by a residual addition, as
\begin{equation}\label{eq:resblock}
R_i(H^{i-1};W_R^i) = \tau( f(\tau(f(H^{i-1};W_R^{i,1}));W_R^{i,2}) + H^{i-1}).
\end{equation}

Here, $H^i$ is the output of the $i$-th residual block, $W_R^i$ is the parameter set of the residual block, and $W_R^{i,j}$ is the parameter of the $j$-th convolution layer in the $i$-th block. With this notation, we denote the output feature of the final residual block of ResNet as $H^u$, which becomes the input to the upsampling block.
\begin{equation}\label{eq:resnet}
H^u = R_u\left(\dots\left(R_1\left(f\left(\boldsymbol{X};W_c\right);W_R^1\right)\right)\dots;W_R^u\right).
\end{equation}

Note that because our model has a single convolution layer before each residual block, the first residual block gets $f(\boldsymbol{X}; W_c)$ as input, where $W_c$ is the parameter of the convolution layer.

In contrast to ResNet, our CARN model has a local cascading block illustrated in block (c) of Fig.~\ref{fig:block} instead of a plain residual block. In here, we denote $B^{i, j}$ as the output of the $j$-th residual block in the $i$-th cascading block, and $W_c^i$ as the set of parameters of the $i$-th local cascading block. Then, the $i$-th local cascading block $B_{local}^{i}$ is defined as
\begin{equation}\label{eq:local_carn}
B_{local}^i\left(H^{i-1};W_l^i\right) \equiv B^{i, U},
\end{equation}
where $B^{i,U}$ is defined recursively from the $B^{i,u}$'s as:
\begin{align*}
B^{i,0} &= H^{i-1} \nonumber \\
B^{i,u} &= f\left(\left[I, B^{i,0},\dots,B^{i,u-1},R^{u}\left(B^{i,u-1};W_R^{u}\right)\right];W_c^{i,u}\right) \quad\text{for $u=1,\dots,U$.}
\end{align*}

Finally, we can define the output feature of the final cascading block $H^b$ by combining both the local and global cascading. Here, $H^0$ is the output of the first convolution layer. And we we fix $u=b=3$ for our CARN and CARN-M.
\begin{equation}\label{eq:carn}
\begin{split}
H^0 &= f\left(\boldsymbol{X};W_c\right) \\
H^b &= f\left(\left[H^0,\dots,H^{b-1},B^u_{local}\right(H^{b-1};W_B^b\left)\right]\right)\quad\text{for $b=1,\dots,B$.}
\end{split}
\end{equation}

The main difference between CARN and ResNet lies in the cascading mechanism. As shown in Fig.~\ref{fig:model}, CARN has global cascading connections represented as the blue arrows, each of which is followed by a 1$\times$1 convolution layer. Cascading on both the local and global levels has two advantages: \textbf{1)} The model incorporates features from multiple layers, which allows learning multi-level representations. \textbf{2)} Multi-level cascading connection behaves as multi-level shortcut connections that quickly propagate information from lower to higher layers (and vice-versa, in case of back-propagation).

CARN adopts a multi-level representation scheme as in ~\cite{lee2017multi,long2015fully}, but we apply this arrangement to a variety of feature levels to boost performance, as shown in equation~\ref{eq:carn}. By doing so, our model reconstructs the LR image based on multi-level features. This facilitates the model to restore the details and contexts of the image simultaneously. As a result, our models effectively improve not only primitive objects but also complex objects.

Another reason for adopting the cascading scheme is two-fold:
First, the propagation of information follows multiple paths~\cite{densenet,unet}.
Second, by adding extra convolution layers, our model can learn to choose the right pathway with the given input information flows. However, the strength of multiple shortcuts is degraded when we use only one of local or global cascading, especially the local connection. We elaborate the details and present a case study on the effects of cascading mechanism in Section~\ref{subsec:analysis}.

\subsection{Efficient Cascading Residual Network}
\label{subsec:carn-m}
To improve the efficiency of CARN, we propose an efficient residual (residual-E) block. We use a similar approach to the MobileNet~\cite{mobilenets}, but use group convolution instead of depthwise convolution. Our residual-E block consists of two 3$\times$3 group and one pointwise convolution, as shown in Fig.~\ref{fig:block} (b). The advantage of using group convolution over the depthwise convolution is that it makes the efficiency of the model tunable. The user can choose the group size appropriately since the group size and performance are in a trade-off relationship. The analysis on the cost efficiency of using the residual-E block is as follows.

Let $K$ be the kernel size and $C_{in}, C_{out}$ be the number of input and output channels. Because we retain the feature resolution of the input and output by padding, we can denote $F$ to be both the input and output feature size. Then, the cost of a plain residual block is as $2\times\left(K \cdot K \cdot C_{in} \cdot C_{out} \cdot F \cdot F\right)$. Note that we only count the cost of convolution layers and ignore the addition or activation because both the plain and the efficient residual blocks have the same amount of cost in terms of addition and activation.

Let $G$ be the group size. Then, the cost of a residual-E block, which consist of two group convolutions and one pointwise convolution, is as given in equation~\ref{eq:reseblock}.

\begin{equation} \label{eq:reseblock}
2\times\left(K \cdot K \cdot C_{in} \cdot \frac{C_{out}}{G} \cdot F \cdot F\right) + C_{in} \cdot C_{out} \cdot F \cdot F
\end{equation}

By changing the plain residual block to our efficient residual block, we can reduce the computation by the ratio of

\begin{equation}
\frac{2\times\left(K \cdot K \cdot C_{in} \cdot \frac{C_{out}}{G} \cdot F \cdot F\right) + C_{in} \cdot C_{out} \cdot F \cdot F}{2\times\left(K \cdot K \cdot C_{in} \cdot C_{out} \cdot F \cdot F\right)} \\
= \frac{1}{G} + \frac{1}{2K^2}.
\end{equation}

Because our model uses a kernel of size 3$\times$3 for all group convolutions, and the number of channels is constantly 64, using an efficient residual block instead of a standard residual block can reduce the computation from 1.8 up to 14 times depending on the group size. To find the best trade-off between performance and computation, we performed an extensive case study in Section~\ref{subsec:analysis}.

To further reduce the parameters, we apply a technique similar to the one used by the recursive network. That is, we make the parameters of the Cascading blocks shared, effectively making the blocks recursive. Fig.~\ref{fig:block}  (d) shows our block after applying the recursive scheme. This approach reduces the parameters by up to three times of their original number.

\subsection{Comparison to Recent Models}
\textbf{Comparison to SRDenseNet.} SRDenseNet~\cite{srdense} uses dense block and skip connection. The differences from our model are: \textbf{1)} We use global cascading, which is more general than the skip connection. In SRDenseNet, all levels of features are combined at the end of the final dense block, but our global cascading scheme connects all blocks, which behaves as multi-level skip connection. \textbf{2)} SRDenseNet preserves local information of dense block via concatenation operations, while we gather it progressively by $1\times1$ convolution layers. The use of additional 1$\times$1 convolution layers results in a higher representation power.

\noindent\textbf{Comparison to MemNet.} The motivation of MemNet~\cite{memnet} and ours is similar. However, there are two main differences from our mechanism. \textbf{1)} Inside of the memory blocks of MemNet, the output features of each recursive units are concatenated at the end of the network and then fused with $1\times1$ convolution. On the other hand, we fuse the features at every possible point in the local block, which can boost up the representation power via the additional convolution layers and nonlinearity. In general, this representation power is often not met because of the difficulty of training. However, we overcome this problem by using both local and global cascading mechanisms. We will discuss the details on Section~\ref{subsec:analysis}.
\textbf{2)} MemNet takes upsampled images as input so the number of multi-adds is larger than ours. The input to our model is a LR image and we upsample it at the end of the network in order to achieve computational efficiency.

\section{Experimental Results}
\label{sec:experiments}

\subsection{Datasets}
\label{subsec:dataset}
There exist diverse single image super-resolution datasets, but the most widely used ones are the 291 image set by Yang et al.~\cite{yang2010} and the Berkeley Segmentation Dataset~\cite{bsd2011}. However, because these two do not have sufficient images for training a deep neural network, we additionally use the DIV2K dataset~\cite{div2k}. The DIV2K dataset is a newly-proposed high-quality image dataset, which consists of 800 training images, 100 validation images, and 100 test images. Because of the richness of this dataset, recent SR models~\cite{mdsr2017,btsrn2017,cnf2017,selnet} use DIV2K as well. We use the standard benchmark datasets such as Set5~\cite{set5}, Set14~\cite{yang2010}, B100~\cite{b100} and Urban100~\cite{urban100} for testing and benchmarking.

\subsection{Implementation and Training Details}
\label{subsec:implementation}
We use the RGB input patches of size 64$\times$64 from the LR images for training. We sample the LR patches randomly and augment them with random horizontal flips and 90 degree rotation. We train our models with the ADAM optimizer~\cite{adam} by setting $\beta_{1} = 0.9$, $\beta_{2} = 0.999$, and $\epsilon = 10^{-8}$ in $6 \times 10^5$ steps. The minibatch size is 64, and the learning rate begins with $10^{-4}$ and is halved every $4 \times 10^5$ steps. All the weights and biases are initialized by $\theta \sim U(-k,\ k)$ with $k = 1/\sqrt{c_{in}}$ where, $c_{in}$ is the number of channels of input feature map.

The most well-known and effective weight initialization methods are given by Glorot et al.~\cite{glorot2010} and He et al.~\cite{he2015delving}. However, such initialization routines tend to set the weights of our multiple narrow 1$\times$1 convolution layers very high, resulting in an unstable training. Therefore, we sample the initial values from a uniform distribution to alleviate the initialization problem.

\begin{figure}[t]
    \centering
    \includegraphics[width=\textwidth]{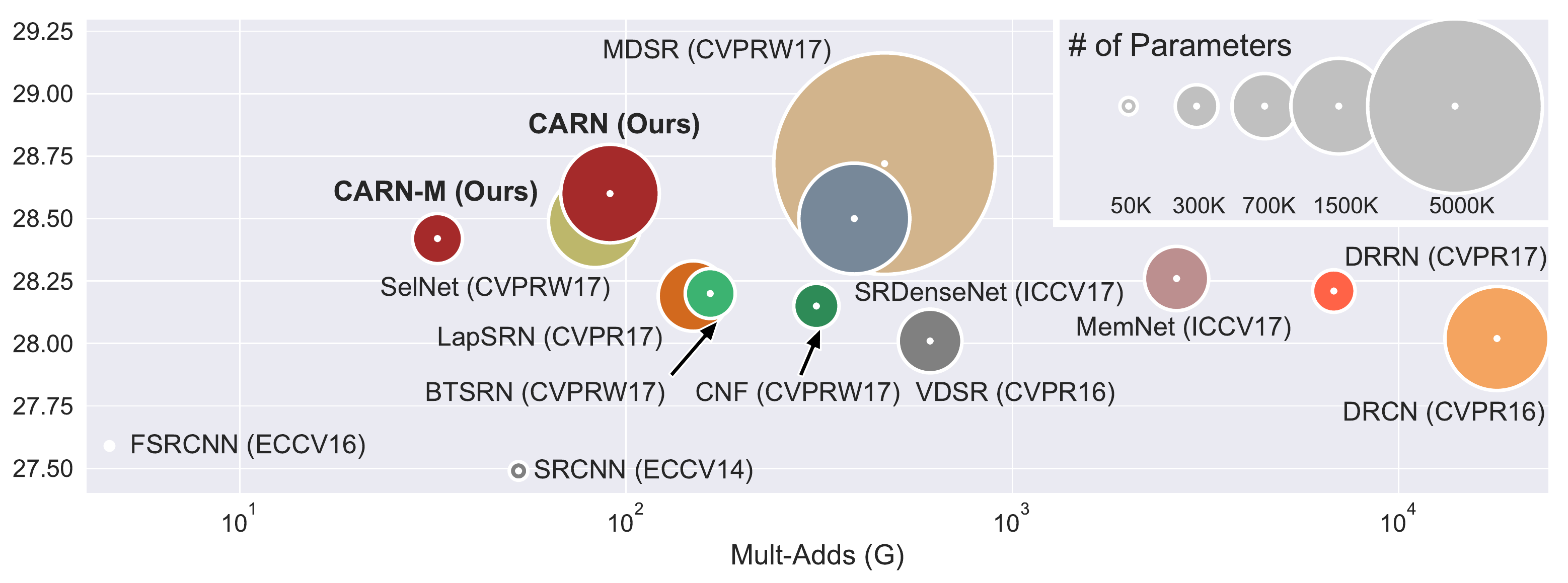}
    \caption{Trade-off between performance vs. number of operations and parameters on Set14 $\times$4 dataset. The $x$-axis and the $y$-axis denote the Multi-Adds and PSNR, and the size of the circle represents the number of parameters. The Mult-Adds is computed by assuming that the resolution of HR image is 720p.}
    \label{fig:benchmark}
\end{figure}

To train our model in a multi-scale manner, we first set the scaling factor to one of $\times$2, $\times$3, and $\times$4 because our model can only process a single scale for each batch. Then, we construct and argument our input batch, as described above. We use the L1 loss as our loss function instead of the L2. The L2 loss is widely used in the image restoration task due to its relationship with the peak signal-to-noise ratio (PSNR). However, in our experiments, L1 provides better convergence and performance. The downside of the L1 loss is that the convergence speed is relatively slower than that of L2 without the residual block. However, this drawback could be mitigated by using a ResNet style model.

\subsection{Comparison with State-of-the-art Methods}
\label{subsec:comparison}
We compare the proposed CARN and CARN-M with state-of-the-art SR methods on two commonly-used image quality metrics: PSNR and the structural similarity index (SSIM)~\cite{ssim}. One thing to note here is that we represent the number of operations by Mult-Adds. Mult-Adds is the number of composite multiply-accumulate operations for a single image. We assume the HR image size to be 720p (1280$\times$720) to calculate Multi-Adds. In Fig.~\ref{fig:benchmark}, we compare our CARN family against the various benchmark algorithms in terms of the Mult-Adds and the number of the parameters on the Set14 $\times$4 dataset. Here, our CARN model outperforms all state-of-the-art models that have less than 5M parameters. Especially, CARN has similar number of parameters to that of DRCN~\cite{drcn2016}, SelNet~\cite{selnet} and SRDenseNet~\cite{srdense}, but we outperform all three models. The MDSR~\cite{mdsr2017} achieves better performance than ours, which is not surprising because MDSR has 8M parameters which are nearly six times more parameters than ours. The CARN-M model also outperforms most of the benchmark methods and shows comparable results against the heavy models.

Moreover, our models are most efficient in terms of the computation cost: CARN shows second best results with 90.9G Mult-Adds, which is on par with SelNet~\cite{selnet}. This efficiency mainly comes from the \textit{late-upsample} approach that many recent models~\cite{srdense,lapsrn2017,fsrcnn2016} used. In addition, our novel cascading mechanism shows increased performance compared to other similar approaches. For example, CARN outperforms its most similar model SelNet by a margin of 0.11 PSNR using similar number of operations.
Also, the CARN-M model obtains comparable results against computationally-expensive models, while only requiring the similar number of the operations with respect to SRCNN.

\begin{table}[t]
\scriptsize
\setlength{\tabcolsep}{2pt}
\caption{Quantitative results of deep learning-based SR algorithms. Red/blue text: best/second-best.}
\begin{center}
\begin{tabular}{c l r r c c c c}
\hline\Tstrut
\multirow{2}{*}{Scale} & \multirow{2}{*}{Model} & \multirow{2}{*}{Params} & \multirow{2}{*}{MultAdds} & Set5 & Set14 & B100 & Urban100 \\
&  &  & & PSNR/SSIM & PSNR/SSIM & PSNR/SSIM & PSNR/SSIM
\Bstrut\\\hline\Tstrut
\multirow{12}{*}{2} & SRCNN\cite{srcnn2014}
                           & 57K    & 52.7G    & 36.66/0.9542 & 32.42/0.9063 & 31.36/0.8879 & 29.50/0.8946 \\
& FSRCNN\cite{fsrcnn2016}  & 12K    & 6.0G     & 37.00/0.9558 & 32.63/0.9088 & 31.53/0.8920 & 29.88/0.9020 \\
& VDSR\cite{vdsr2016}      & 665K   & 612.6G   & 37.53/0.9587 & 33.03/0.9124 & 31.90/0.8960 & 30.76/0.9140 \\
& DRCN\cite{drcn2016}      & 1,774K & 17,974.3G & 37.63/0.9588 & 33.04/0.9118 & 31.85/0.8942 & 30.75/0.9133 \\
& CNF\cite{cnf2017}        & 337K   & 311.0G   & 37.66/0.9590 & 33.38/0.9136 & 31.91/0.8962 & -\\
& LapSRN\cite{lapsrn2017}  & 813K   & 29.9G    & 37.52/0.9590 & 33.08/0.9130 & 31.80/0.8950 & 30.41/0.9100 \\
& DRRN\cite{drnn2017}      & 297K   & 6,796.9G & 37.74/0.9591 & 33.23/0.9136 & 32.05/0.8973 & 31.23/0.9188 \\
& BTSRN\cite{btsrn2017}    & 410K   & 207.7G   & 37.75/-\HSE  & 33.20/-\HSE  & 32.05/-\HSE  & \blue{31.63}/-\HSE \\
& MemNet\cite{memnet}      & 677K   & 2,662.4G   & \blue{37.78}/\blue{0.9597} & 33.28/0.9142 & \blue{32.08}/\blue{0.8978} & 31.31/\blue{0.9195} \\
& SelNet\cite{selnet}      & 974K   & 225.7G   & \red{37.89}/\red{0.9598}  & \red{33.61}/\blue{0.9160}  & \blue{32.08}/\red{0.8984} & - \\
& CARN (ours)              & 1,592K & 222.8G   & 37.76/0.9590 & \blue{33.52}/\red{0.9166} & \red{32.09}/\blue{0.8978}	& \red{31.92}/\red{0.9256} \\
& CARN-M (ours)            & 412K   & 91.2G    & 37.53/0.9583 & 33.26/0.9141 & 31.92/0.8960	& 31.23/0.9193\Bstrut \\\hline\Tstrut
\multirow{11}{*}{3} & SRCNN\cite{srcnn2014}
                           & 57K    & 52.7G    & 32.75/0.9090 & 29.28/0.8209 & 28.41/0.7863 & 26.24/0.7989 \\
& FSRCNN\cite{fsrcnn2016}  & 12K    & 5.0G     & 33.16/0.9140 & 29.43/0.8242 & 28.53/0.7910 & 26.43/0.8080 \\
& VDSR\cite{vdsr2016}      & 665K   & 612.6G   & 33.66/0.9213 & 29.77/0.8314 & 28.82/0.7976 & 27.14/0.8279 \\
& DRCN\cite{drcn2016}      & 1,774K & 17,974.3G & 33.82/0.9226 & 29.76/0.8311 & 28.80/0.7963 & 27.15/0.8276 \\
& CNF\cite{cnf2017}        & 337K   & 311.0G   & 33.74/0.9226 & 29.90/0.8322 & 28.82/0.7980 & -\\
& DRRN\cite{drnn2017}      & 297K   & 6,796.9G & 34.03/0.9244 & 29.96/0.8349 & 28.95/0.8004 & 27.53/0.8378 \\
& BTSRN\cite{btsrn2017}    & 410K   & 176.2G   & 34.03/-\HSE  & 29.90/-\HSE  & \blue{28.97}/-\HSE & \blue{27.75}/-\HSE \\
& MemNet\cite{memnet}      & 677K   & 2,662.4G   & 34.09/0.9248 & 30.00/0.8350 & 28.96/0.8001 & 27.56/0.8376 \\
& SelNet\cite{selnet}      & 1,159K & 120.0G   & \blue{34.27}/\red{0.9257}  & \red{30.30}/\blue{0.8399}  & \blue{28.97}/\blue{0.8025} & - \\
& CARN (ours)              & 1,592K & 118.8G   & \red{34.29}/\blue{0.9255} & \blue{30.29}/\red{0.8407} & \red{29.06}/\red{0.8034}	& \red{28.06}/\red{0.8493} \\
& CARN-M (ours)            & 412K   & 46.1G    & 33.99/0.9236 & 30.08/0.8367 & 28.91/0.8000	& 27.55/\blue{0.8385}\Bstrut \\\hline\Tstrut
\multirow{13}{*}{4} &SRCNN\cite{srcnn2014}
                           & 57K    & 52.7G    & 30.48/0.8628 & 27.49/0.7503 & 26.90/0.7101 & 24.52/0.7221 \\
& FSRCNN\cite{fsrcnn2016}  & 12K    & 4.6G     & 30.71/0.8657 & 27.59/0.7535 & 26.98/0.7150 & 24.62/0.7280 \\
& VDSR\cite{vdsr2016}      & 665K   & 612.6G   & 31.35/0.8838 & 28.01/0.7674 & 27.29/0.7251 & 25.18/0.7524 \\
& DRCN\cite{drcn2016}      & 1,774K & 17,974.3G & 31.53/0.8854 & 28.02/0.7670 & 27.23/0.7233 & 25.14/0.7510 \\
& CNF\cite{cnf2017}        & 337K   & 311.0G   & 31.55/0.8856 & 28.15/0.7680 & 27.32/0.7253 & -\\
& LapSRN\cite{lapsrn2017}  & 813K   & 149.4G   & 31.54/0.8850 & 28.19/0.7720 & 27.32/0.7280 & 25.21/0.7560 \\
& DRRN\cite{drnn2017}      & 297K   & 6,796.9G & 31.68/0.8888 & 28.21/0.7720 & 27.38/0.7284 & 25.44/0.7638\\
& BTSRN\cite{btsrn2017}    & 410K   & 165.2G   & 31.85/-\HSE  & 28.20/-\HSE  & 27.47/-\HSE  & 25.74/-\HSE \\
& MemNet\cite{memnet}      & 677K   & 2,662.4G   & 31.74/0.8893 & 28.26/0.7723 & 27.40/0.7281 & 25.50/0.7630 \\
& SelNet\cite{selnet}      & 1,417K & 83.1G    & 32.00/0.8931 & 28.49/\blue{0.7783} & 27.44/0.7325 & - \\
& SRDenseNet\cite{srdense} & 2,015K & 389.9G   & \blue{32.02}/\blue{0.8934} & \blue{28.50}/0.7782 & \blue{27.53}/\blue{0.7337} & \blue{26.05}/\blue{0.7819} \\
& CARN (ours)              & 1,592K & 90.9G    & \red{32.13}/\red{0.8937} & \red{28.60}/\red{0.7806} & \red{27.58}/\red{0.7349}	& \red{26.07}/\red{0.7837} \\
& CARN-M (ours)            & 412K   & 32.5G    & 31.92/0.8903 & 28.42/0.7762 & 27.44/0.7304	& 25.62/0.7694\Bstrut\\\hline
\end{tabular}
\end{center}
\label{table:benchmark}
\end{table}

Table~\ref{table:benchmark} also shows the quantitative comparisons of the performances over the benchmark datasets. Note that MDSR is excluded from this table, because we only compare models that have roughly similar number of parameters as ours; MDSR has a parameter set whose size is four times larger than that of the second-largest model. Our CARN exceeds all the previous methods on numerous benchmark dataset.
CARN-M model achieves comparable results using very few operations. We would also like to emphasize that although CARN-M has more parameters than SRCNN or DRRN, it is tolerable in real-world scenarios. The sizes of SRCNN and CARN-M are 200KB and 1.6MB, respectively, all of which are acceptable on recent mobile devices.

To make our models even more lightweight, we apply the multi-scale learning approach. The benefit of using multi-scale learning is that it can process multiple scales using a single trained model. 
This helps us alleviate the burden of heavy-weight model size when deploying the SR application on mobile devices; CARN(-M) only needs a single fixed model for multiple scales, whereas even the state-of-the-art algorithms require to train separate models for each supported scale. This property is well-suited for real-world products because the size of the applications has to be fixed while the scale of given LR images could vary. Using the multi-scale learning to our models increases the number of parameters, since the network has to contain possible upsampling layers. On the other hand, VDSR and DRRN do not require this extra burden, even if multi-scale learning is performed, because they upsample the image before processing it.

\begin{figure*}[htbp]
\centering
\includegraphics[height=0.95\textheight]{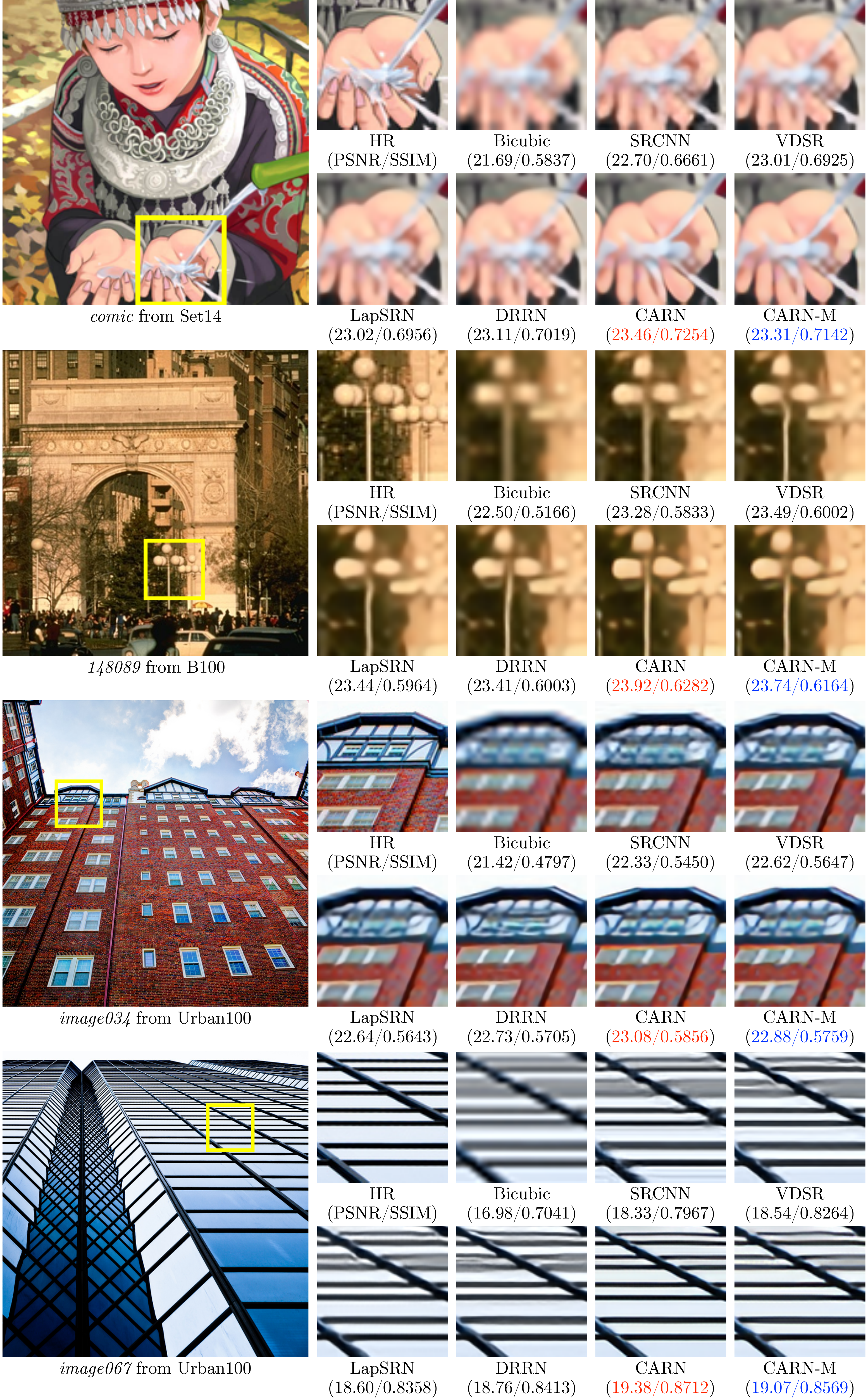}
\caption{Visual qualitative comparison on $\times$4 scale datasets.}
\label{fig:comparison}
\end{figure*}

\begingroup
\renewcommand{\arraystretch}{1.1}
\setlength{\tabcolsep}{5pt}
\begin{table}[tbp]
\caption{Effects of the global and local cascading modules measured on the Set14 $\times$4 dataset. CARN-NL represents CARN without local cascading and CARN-NG without global cascading. CARN is our final model.}
\begin{center}
\begin{tabular}{| c | c c c c |}
\hline
                 & Baseline & CARN-NL   & CARN-NG          & CARN \\\hline
Local Cascading  &          &            & \Checkmark & \Checkmark \\
Global Cascading &          & \Checkmark &            & \Checkmark \\\hline
\# Params.       & 1,444K   & 1,481K     & 1,555K     & 1,592K \\
PSNR             & 28.43    & 28.45      & 28.42      & \textbf{28.52}
\\\hline
\end{tabular}
\end{center}
\label{table:ablation1}
\end{table}
\endgroup
In Fig.~\ref{fig:comparison}, we visually illustrate the qualitative comparisons over three datasets (Set14, B100 and Urban100) for $\times$4 scale. It can be seen that our model works better than others and accurately reconstructs not only stripes and line patterns, but also complex objects such as hand and street lamps.

\subsection{Model Analysis}
\label{subsec:analysis}
To further investigate the performance behavior of the proposed methods, we analyze our models via ablation study. First, we show how local and global cascading modules affect the performance of CARN. Next, we analyze the trade-off between performance vs. parameters and operations.

\subsubsection{Cascading Modules.}
Table \ref{table:ablation1} presents the ablation study on the effect of local and global cascading modules. In this table, the baseline is ResNet, CARN-NL is CARN without local cascading and CARN-NG is CARN without global cascading. The network topologies are all same, but because of the 1$\times$1 convolution layer, the overall number of parameters is increased by up to 10\%.

We see that the model with only global cascading (CARN-NL) shows better performance than the baseline because the global cascading mechanism effectively carries mid- to high-level frequency signals from shallow to deep layers. Furthermore, by gathering all features before the upsampling layers, the model can better leverage multi-level representations. By incorporating multi-level representations, the CARN model can consider a variety of information from many different receptive fields when reconstructing the image.

Somewhat surprisingly, using only local cascading blocks (CARN-NG) harms the performance. As discussed in He et al.~\cite{he2016identity}, multiplicative manipulations such as 1$\times$1 convolution on the shortcut connection can hamper information propagation, and thus lead to complications during optimization. Similarly, cascading connections in the local cascading blocks of CARN-NG behave as shortcut connections inside the residual blocks. Because these connections consist of concatenation and 1$\times$1 convolutions, it is natural to expect performance degradation. That is, the advantage of multi-level representation is limited to the inside of each local cascading block. Therefore, there appears to be no benefit of using the cascading connection because of the increased number of multiplication operations in the cascading connection. However, CARN uses both local and global cascading levels and outperforms all three models. This is because the global cascading mechanism eases the information propagation issues that CARN-NG suffers from. In detail, information propagates globally via global cascading, and information flows in the local cascading blocks are fused with the ones that come through global connections. By doing so, information is transmitted by multiple shortcuts and thus mitigates the vanishing gradient problem. In other words, the advantage of multi-level representation is leveraged by the global cascading connections, which help the information to propagate to higher layers.

\subsubsection{Efficiency Trade-off.}
\begin{figure}[tbp]
	\centering
	\begin{subfigure}[b]{0.48\textwidth}
		\includegraphics[width=\textwidth]{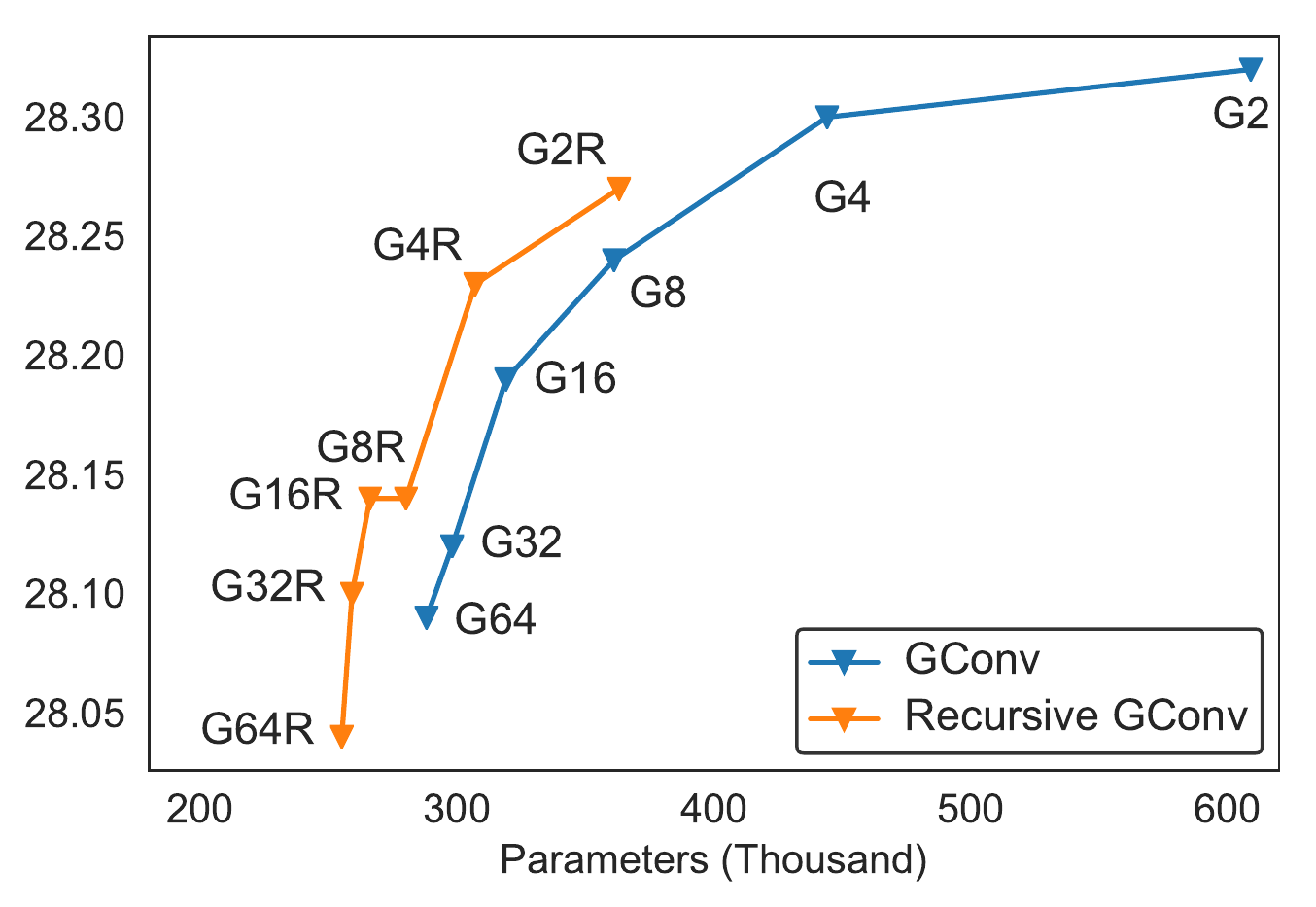}
		\caption{Trade-off of parameters-PSNR}
		\label{fig:params}
	\end{subfigure}
	\begin{subfigure}[b]{0.48\textwidth}
		\includegraphics[width=\textwidth]{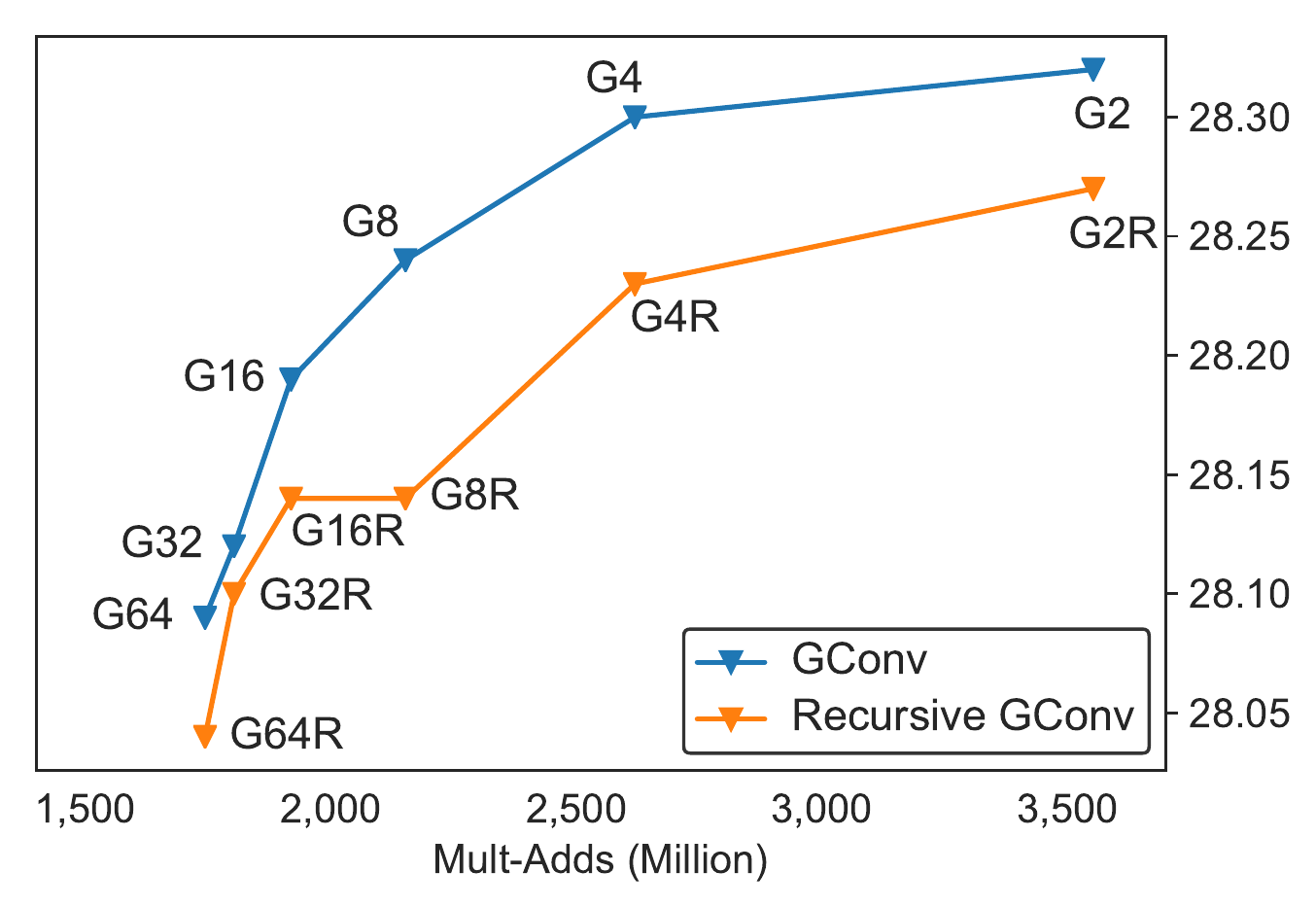}
		\caption{Trade-off of operations-PSNR}
		\label{fig:macs}
	\end{subfigure}
	\caption{Results of using efficient residual block and recursive network in terms of PSNR vs. parameters(\textbf{left}) and PSNR vs. operations(\textbf{right}). We evaluate all models on Set14 with $\times$4 scale. \textit{GConv} represents the group size of group convolution and \textit{R} means the model with the recursive network scheme (i.e., \textit{G4R} represents group four with recursive cascading blocks).}
	\label{fig:efficient}
\end{figure}

Fig.~\ref{fig:efficient} depicts the trade-off study of PSNR vs. parameters, and PSNR vs. operations in relation to the efficient residual block and recursive network. In this experiment, we evaluate all possible group sizes of the efficient residual block for both the recursive and non-recursive cases. In both graphs, the blue line represents the model that does not use the recursive scheme and the orange line is the model that uses recursive cascading block.

Although all efficient models perform worse than the CARN, which shows 28.70 PSNR, the number of parameters and operations are decreased dramatically. For example, the \textit{G64} shows a five-times reduction in both parameters and operations. However, unlike the comparable result that is shown in Howard et al.~\cite{mobilenets}, the degradation of performance is more pronounced in our case.

Next, we observe the case which uses the recursive scheme. As illustrated in Fig.~\ref{fig:macs}, there is no change in the Mult-Adds but the performance worsens, which seems reasonable given the decreased number of parameters in the recursive scheme. On the other hand, Fig.~\ref{fig:params} shows that using the recursive scheme makes the model achieve better performance with fewer parameters. Based on these observations, we decide to choose the group size as four in the efficient residual block and use the recursive network scheme as our CARN-M model. By doing so, CARN-M reduces the number of parameters by five times and the number of operations by nearly four times with a loss of 0.29 PSNR compared to CARN.

\section{Conclusion}
\label{sec:conclusion}
In this work, we proposed a novel cascading network architecture that can perform SISR accurately and efficiently. The main idea behind our architecture is to add multiple cascading connections starting from each intermediary layer to the others. Such connections are made on both the local (block-wise) and global (layer-wise) levels, which allows for the efficient flow of information and gradient. Our experiments show that employing both types of connections greatly outperforms those using only one or none at all.

We wish to further develop this work by applying our technique to video data. Many streaming services require large storage to provide high-quality videos. In conjunction with our approach, one may devise a service that stores low-quality videos that go through our SR system to produce high-quality videos on-the-fly.

\subsection*{Acknowledgement.}
This research was supported through the National Research Foundation of Korea (NRF) funded by the Ministry of Education: NRF-2016R1D1A1B03933875 (K.-A. Sohn) and NRF-2016R1A6A3A11932796 (B. Kang).

\bibliographystyle{splncs04}
\bibliography{}

\begin{thebibliography}{10}
\providecommand{\url}[1]{\texttt{#1}}
\providecommand{\urlprefix}{URL }
\providecommand{\doi}[1]{https://doi.org/#1}

\bibitem{div2k}
Agustsson, E., Timofte, R.: Ntire 2017 challenge on single image
  super-resolution: Dataset and study. In: Proceedings of the Conference on
  Computer Vision and Pattern Recognition (CVPR) Workshops (2017)

\bibitem{bsd2011}
Arbelaez, P., Maire, M., Fowlkes, C., Malik, J.: Contour detection and
  hierarchical image segmentation. IEEE transactions on pattern analysis and
  machine intelligence  \textbf{33}(5),  898--916 (2011)

\bibitem{set5}
Bevilacqua, M., Roumy, A., Guillemot, C., Alberi{-}Morel, M.: Low-complexity
  single-image super-resolution based on nonnegative neighbor embedding. In:
  Proceedings of the British Machine Vision Conference (BMVC) (2012)

\bibitem{selnet}
Choi, J.S., Kim, M.: A deep convolutional neural network with selection units
  for super-resolution. In: Proceedings of the Conference on Computer Vision
  and Pattern Recognition (CVPR) Workshops (2017)

\bibitem{imagenet2009}
Deng, J., Dong, W., Socher, R., Li, L.J., Li, K., Fei-Fei, L.: Imagenet: A
  large-scale hierarchical image database. In: Proceedings of the Conference on
  Computer Vision and Pattern Recognition (CVPR) (2009)

\bibitem{srcnn2014}
Dong, C., Loy, C.C., He, K., Tang, X.: Learning a deep convolutional network
  for image super-resolution. In: Proceedings of the European Conference on
  Computer Vision (ECCV) (2014)

\bibitem{fsrcnn2016}
Dong, C., Loy, C.C., Tang, X.: Accelerating the super-resolution convolutional
  neural network. In: Proceedings of the European Conference on Computer Vision
  (ECCV) (2016)

\bibitem{btsrn2017}
Fan, Y., Shi, H., Yu, J., Liu, D., Han, W., Yu, H., Wang, Z., Wang, X., Huang,
  T.S.: Balanced two-stage residual networks for image super-resolution. In:
  Proceedings of the Conference on Computer Vision and Pattern Recognition
  (CVPR) Workshops (2017)

\bibitem{fasterrcnn}
Girshick, R.: Fast r-cnn. In: Proceedings of the International Conference on
  Computer Vision (ICCV) (2015)

\bibitem{glorot2010}
Glorot, X., Bengio, Y.: Understanding the difficulty of training deep
  feedforward neural networks. In: Proceedings of the International Conference
  on Artificial Intelligence and Statistics (2010)

\bibitem{han2015deep}
Han, S., Mao, H., Dally, W.J.: Deep compression: Compressing deep neural
  networks with pruning, trained quantization and huffman coding. Proceedings
  of the International Conference on Learning Representations (ICLR)  (2016)

\bibitem{he2015delving}
He, K., Zhang, X., Ren, S., Sun, J.: Delving deep into rectifiers: Surpassing
  human-level performance on imagenet classification. In: Proceedings of the
  International Conference on Computer Vision (ICCV) (2015)

\bibitem{resnet}
He, K., Zhang, X., Ren, S., Sun, J.: Deep residual learning for image
  recognition. In: Proceedings of the Conference on Computer Vision and Pattern
  Recognition (CVPR) (2016)

\bibitem{resnet2016}
He, K., Zhang, X., Ren, S., Sun, J.: Deep residual learning for image
  recognition. In: Proceedings of the Conference on Computer Vision and Pattern
  Recognition (CVPR) (2016)

\bibitem{he2016identity}
He, K., Zhang, X., Ren, S., Sun, J.: Identity mappings in deep residual
  networks. In: Proceedings of the European Conference on Computer Vision
  (ECCV) (2016)

\bibitem{mobilenets}
Howard, A.G., Zhu, M., Chen, B., Kalenichenko, D., Wang, W., Weyand, T.,
  Andreetto, M., Adam, H.: Mobilenets: Efficient convolutional neural networks
  for mobile vision applications. arXiv preprint arXiv:1704.04861  (2017)

\bibitem{densenet}
Huang, G., Liu, Z., van~der Maaten, L., Weinberger, K.Q.: Densely connected
  convolutional networks. In: Proceedings of the Conference on Computer Vision
  and Pattern Recognition (CVPR) (2017)

\bibitem{urban100}
Huang, J.B., Singh, A., Ahuja, N.: Single image super-resolution from
  transformed self-exemplars. In: Proceedings of the Conference on Computer
  Vision and Pattern Recognition (CVPR) (2015)

\bibitem{squeezenet}
Iandola, F.N., Han, S., Moskewicz, M.W., Ashraf, K., Dally, W.J., Keutzer, K.:
  Squeezenet: Alexnet-level accuracy with 50x fewer parameters and< 0.5 mb
  model size. arXiv preprint arXiv:1602.07360  (2016)

\bibitem{vdsr2016}
Kim, J., Kwon~Lee, J., Mu~Lee, K.: Accurate image super-resolution using very
  deep convolutional networks. In: Proceedings of the Conference on Computer
  Vision and Pattern Recognition (CVPR) (2016)

\bibitem{drcn2016}
Kim, J., Kwon~Lee, J., Mu~Lee, K.: Deeply-recursive convolutional network for
  image super-resolution. In: Proceedings of the Conference on Computer Vision
  and Pattern Recognition (CVPR) (2016)

\bibitem{adam}
Kingma, D.P., Ba, J.: Adam: A method for stochastic optimization. Proceedings
  of the International Conference on Learning Representations (ICLR)  (2015)

\bibitem{alexnet}
Krizhevsky, A., Sutskever, I., Hinton, G.E.: Imagenet classification with deep
  convolutional neural networks. In: Proceedings of the Conference on Neural
  Information Processing Systems (NIPS) (2012)

\bibitem{lapsrn2017}
Lai, W.S., Huang, J.B., Ahuja, N., Yang, M.H.: Deep laplacian pyramid networks
  for fast and accurate super-resolution. In: Proceedings of the Conference on
  Computer Vision and Pattern Recognition (CVPR) (2017)

\bibitem{lee2017multi}
Lee, J., Nam, J.: Multi-level and multi-scale feature aggregation using
  pretrained convolutional neural networks for music auto-tagging. IEEE Signal
  Processing Letters  \textbf{24}(8),  1208--1212 (2017)

\bibitem{mdsr2017}
Lim, B., Son, S., Kim, H., Nah, S., Lee, K.M.: Enhanced deep residual networks
  for single image super-resolution. In: Proceedings of the Conference on
  Computer Vision and Pattern Recognition (CVPR) Workshops (2017)

\bibitem{ssd}
Liu, W., Anguelov, D., Erhan, D., Szegedy, C., Reed, S., Fu, C.Y., Berg, A.C.:
  Ssd: Single shot multibox detector. In: Proceedings of the European
  Conference on Computer Vision (ECCV) (2016)

\bibitem{long2015fully}
Long, J., Shelhamer, E., Darrell, T.: Fully convolutional networks for semantic
  segmentation. In: Proceedings of the Conference on Computer Vision and
  Pattern Recognition (CVPR) (2015)

\bibitem{b100}
Martin, D., Fowlkes, C., Tal, D., Malik, J.: A database of human segmented
  natural images and its application to evaluating segmentation algorithms and
  measuring ecological statistics. In: Proceedings of the International
  Conference on Computer Vision (ICCV) (2001)

\bibitem{deconvnet}
Noh, H., Hong, S., Han, B.: Learning deconvolution network for semantic
  segmentation. In: Proceedings of the International Conference on Computer
  Vision (ICCV) (2015)

\bibitem{cnf2017}
Ren, H., El-Khamy, M., Lee, J.: Image super resolution based on fusing multiple
  convolution neural networks. In: Proceedings of the Conference on Computer
  Vision and Pattern Recognition (CVPR) Workshops (2017)

\bibitem{unet}
Ronneberger, O., Fischer, P., Brox, T.: U-net: Convolutional networks for
  biomedical image segmentation. In: Proceedings of the International
  Conference on Medical Image Computing and Computer-Assisted Intervention
  (2015)

\bibitem{espcn2016}
Shi, W., Caballero, J., Husz{\'a}r, F., Totz, J., Aitken, A.P., Bishop, R.,
  Rueckert, D., Wang, Z.: Real-time single image and video super-resolution
  using an efficient sub-pixel convolutional neural network. In: Proceedings of
  the Conference on Computer Vision and Pattern Recognition (CVPR) (2016)

\bibitem{dwconv}
Sifre, L., Mallat, S.: Rigid-motion scattering for image classification. Ph.D.
  thesis, Citeseer (2014)

\bibitem{drnn2017}
Tai, Y., Yang, J., Liu, X.: Image super-resolution via deep recursive residual
  network. In: Proceedings of the Conference on Computer Vision and Pattern
  Recognition (CVPR) (2017)

\bibitem{memnet}
Tai, Y., Yang, J., Liu, X., Xu, C.: Memnet: A persistent memory network for
  image restoration. In: Proceedings of the International Conference on
  Computer Vision (ICCV) (2017)

\bibitem{srdense}
Tong, T., Li, G., Liu, X., Gao, Q.: Image super-resolution using dense skip
  connections. In: Proceedings of the International Conference on Computer
  Vision (ICCV) (2017)

\bibitem{ssim}
Wang, Z., Bovik, A.C., Sheikh, H.R., Simoncelli, E.P.: Image quality
  assessment: from error visibility to structural similarity. IEEE transactions
  on image processing  \textbf{13}(4),  600--612 (2004)

\bibitem{yang2010}
Yang, J., Wright, J., Huang, T.S., Ma, Y.: Image super-resolution via sparse
  representation. IEEE transactions on image processing  \textbf{19}(11),
  2861--2873 (2010)

\bibitem{color}
Zhang, R., Isola, P., Efros, A.A.: Colorful image colorization. In: Proceedings
  of the European Conference on Computer Vision (ECCV) (2016)

\end{thebibliography}
\end{document}